\documentclass{article}
\usepackage{ijcai19}

\usepackage{times}
\usepackage{url}
\usepackage[hidelinks]{hyperref}
\usepackage[utf8]{inputenc}
\usepackage[small]{caption}
\usepackage{graphicx}
\usepackage{subfigure}
\usepackage{color}
\usepackage{booktabs}
\usepackage{epstopdf}
\usepackage{amsmath,amsfonts,amsthm,amssymb}
\newtheorem{example}{Example}

\graphicspath{{./}{fig/}}

\title{MR-GNN: Multi-Resolution and Dual Graph Neural Network for Predicting Structured Entity Interactions}

\author{
Nuo Xu$^1$\and
Pinghui Wang$^{2,1}$\footnote{Corresponding Authors}\thanks{Nuo Xu and Pinghui Wang contributed equally to this work.}\and
Long Chen$^1$\and
Jing Tao$^1$\And
Junzhou Zhao$^{1*}$\\
\affiliations
$^1$MOE NEKEY Lab, Xi'an Jiaotong University, China\\
$^2$Shenzhen Research School, Xi'an Jiaotong University, China\\
\emails
nxu@sei.xjtu.edu.cn,
\{phwang, jtao\}@mail.xjtu.edu.cn,
chenlongche@stu.edu.cn,
junzhouzhao@gmail.com
}

\begin{document}
\maketitle
\setlength{\abovecaptionskip}{0.1cm}
\setlength{\belowcaptionskip}{-0.37cm}
\setlength{\abovedisplayskip}{2pt}
\setlength{\belowdisplayskip}{2pt}

\begin{abstract}
  Predicting interactions between structured entities lies at the core of numerous
tasks such as drug regimen and new material design.
In recent years, graph neural networks have become attractive.
They represent structured entities as graphs, and then extract features from each
individual graph using graph convolution operations.
However, these methods have some limitations: i) their networks only extract
features from a fix-sized subgraph structure (i.e., a fix-sized receptive field)
of each node, and ignore features in substructures of different sizes, and ii)
features are extracted by considering each entity independently, which may not
effectively reflect the interaction between two entities.
To resolve these problems, we present {\em MR-GNN}, an end-to-end graph neural
network with the following features: i) it uses a multi-resolution based
architecture to extract node features from different neighborhoods of each node,
and, ii) it uses dual graph-state long short-term memory networks (LSTMs) to
summarize local features of each graph and extracts the interaction features
between pairwise graphs.
Experiments conducted on real-world datasets show that MR-GNN improves the
prediction of state-of-the-art methods.
\end{abstract}%

\section{Introduction}\label{Intro}

\begin{figure}
  \centering
  \includegraphics[width=0.90\linewidth]{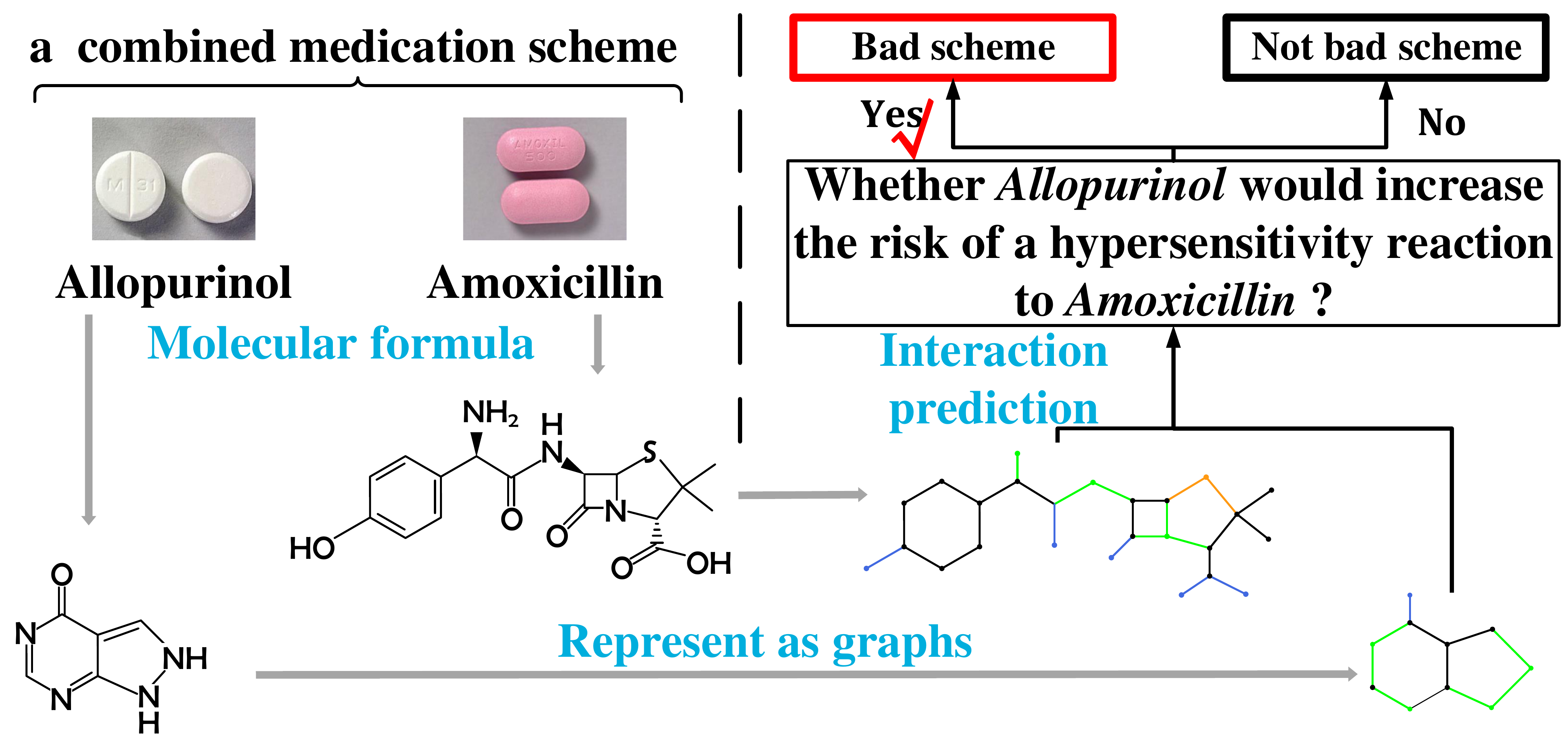}
  \caption{Overview of graph-based framework.
    We transform two drugs {\em Allopurinol} and {\em Amoxicillin} into graphs,
    where nodes represent atoms and edges refer to chemical bonds between atoms,
    and predict interactions between them.
    When there exists an adverse reaction between them, they cannot be taken
    together.
  }
  \label{fig:framework}
\end{figure}

A large variety of applications require understanding the interactions between
structured entities.
For example, when one medicine is taken together with another, each medicine's
intended efficacy may be altered substantially (see Fig.~\ref{fig:framework}).
Understanding their interactions is important to minimize the side effects and
maximize the synergistic benefits~\cite{DeepDDI}.
In chemistry, understanding what chemical reactions will occur between two
chemicals is helpful in designing new materials with desired
properties~\cite{Kwon2017}.
Despite its importance, examining all interactions by performing clinical or
laboratory experiments is impractical due to the potential harms to patients and
also highly time and monetary costs.

Recently, machine learning methods have been proposed to address this problem, and
they are demonstrated to be effective in many
tasks~\cite{Duvenaud2015,Li2017,Tian2016,DeepDDI}.
These methods use features extracted from entities to train a classifier to
predict entity interactions.
However, features have to be carefully provided by domain
experts~\cite{DeepDDI,Tian2016}, and it is labor-intensive.
To automate feature extraction, graph convolution neural networks (GCNs) have been
proposed~\cite{protein_interface,Kwon2017,Zitnik2018}.
GCNs represent structured entities as graphs, and use {\em graph convolution
  operators} to extract features.
One of the state-of-the-art GCN models, proposed by Alex et
al.~\shortcite{protein_interface}, extracts features from the 3-hop neighborhood
of each node.
We thus say that their model uses a fix-sized {\em receptive field (RF)}.
However, using a fix-sized RF to extract features may have limitations,
which can be illustrated by the following example.

\begin{figure}[t]
  \centering
  \includegraphics[width=0.85\linewidth]{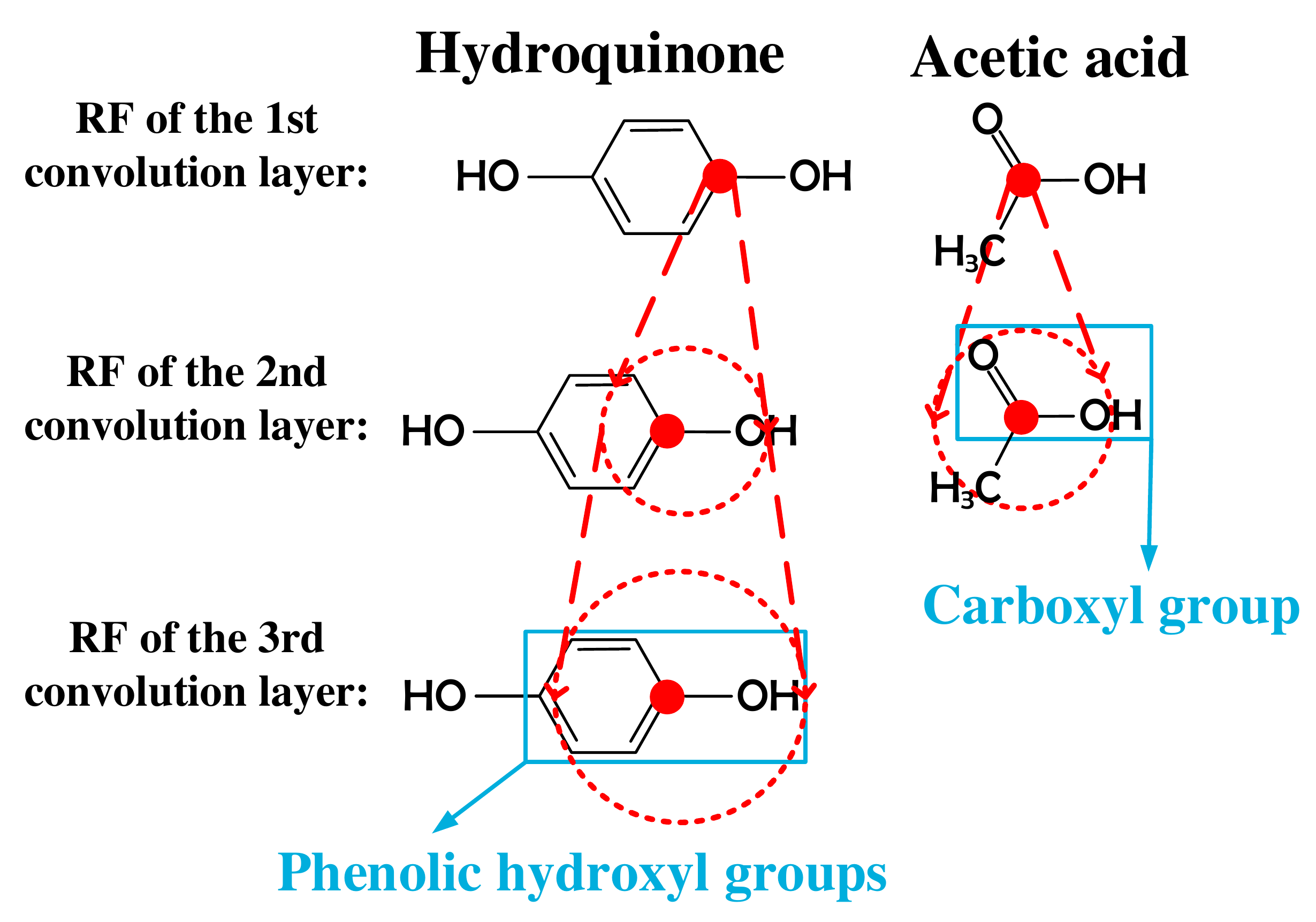}
  \caption{The structure of two weak acids: \emph{Hydroquinone} and \emph{Acetic
      acid}.
    The blue box shows the acidic substructures: ArOH and COOH.
    The red dashed circle shows the receptive field of the corresponding red node
    in different convolution layers.
  }\label{fig:example}
\end{figure}

\begin{example}
  Figure~\ref{fig:example} shows two weak acids, i.e., Hydroquinone and Acetic
  acid.
  They are weak acids due to the existence of substructures phenolic hydroxyl
  (ArOH) and carboxyl (COOH), respectively.
  Representing these two chemical compounds as graphs, we need a three-hop
  neighborhood to accurately extract ArOH from Hydroquinone, and a two-hop
  neighborhood to accurately extract COOH from Acetic acid.
  While using a fix-sized neighborhood will result in that either incomplete
  substructures being extracted (i.e., RF is too small), or useless substructures
  being included (i.e., RF is too large).
\end{example}

Another limitation of existing GCNs is that, they learn each graph's
representation independently, and model the interactions only in the final prediction process.
However, for different entities, the interaction also occurs by substructures of different size.
Take Fig.~2 for example again, when these two weak acids are neutralized with the same strong base, the interaction can be accurately modeled by features of the second convolution layer for \textit{Acetic acid} because the key substructure ArOH can be accurately extracted. But for \textit{Hydroquinone}, the best choice is to model the interaction by features of the third convolution layer. Thus, modeling the interactions only in the final process may make a lot of noise to the prediction.


To address these limitations, this work presents a novel GCN model named
\textbf{M}ulti-\textbf{R}esolution RF based \textbf{G}raph \textbf{N}eural
\textbf{N}etwork (MR-GNN), which leverages different-sized local features and models interaction during the procedure of feature extraction to predict structured entity interactions.

\subsubsection{Overview of our approach.}

MR-GNN uses a multi-resolution RF, which consists of multiple graph convolution
layers with different RFs, to extract local structure features effectively (see
Fig.~2).
When aggregating these multi-resolution local features, MR-GNN uses two key {\em
  dual graph-state LSTMs}.
One is Summary-LSTM (S-LSTM), which aggregates multi-resolution local features for
each graph.
Compared with the straightforward method that simply sums all multi-resolution
features up, S-LSTM learns additional effective features by modeling the diffusion
process of node information in graphs which can greatly enrich the graph
representation.
The other is Interaction-LSTM (I-LSTM), which extracts interaction features
between pairwise graphs during the procedure of feature extraction.

Our contributions are as follows:
\begin{itemize}
\item In MR-GNN, we design a multi-resolution based architecture that mines
  features from multi-scale substructures to predict graph interactions.
  It is more effective than considering only fix-sized RFs.
\item We develop two dual graph-state LSTMs: One summarizes subgraph features of
  multi-sized RFs while modeling the diffusion process of node information, and
  the other extracts interaction features for pairwise graphs during feature
  extraction.
\item Experimental results on two benchmark datasets show that MR-GNN outperforms
  the state-of-the-art methods.
\end{itemize}

\section{Problem Definition}\label{Problem}

\noindent\textbf{Notations.}
We denote a structured entity by a graph $G=(V,E)$, where $V$ is the node set and
$E$ is the edge set.
Each specific node $v_i\in V$ is associated with a $c$-dimension feature vector
$f_i\in\mathbb{R}^c$.
The feature vectors can also be low-dimensional latent representations/embeddings
for nodes or explicit features which intuitively reflects node attributes.
Meanwhile, let $N_i\subseteq V$ denote $v_i$'s neighbors, and $d_i\triangleq
|N_i|$ denote $v_i$'s degree.

\noindent\textbf{Entity Interaction Prediction.}
Let $L\triangleq\{l_i|i=1,2\ldots, k\}$ denote a set of $k$ interaction labels
between two entities.
The entity interaction prediction task is formulated as a supervised learning
problem: Given training dataset $D\triangleq\{(G_X,G_Y)_s,\hat{R}_s\}_{s=1}^q$
where $(G_X,G_Y)_s$ is an input entity pair, and $\hat{R}_s \in L$ is the
corresponding interaction label; let $q$ denote the size of $D$, we want to
accurately predict the interaction label $R \in L$ of an unseen entity pair
$(G_X,G_Y)_\mathrm{new}$.

\section{Method}\label{Method}

In this section, we propose a graph neural network, i.e., MR-GNN, to address the
entity interaction prediction problem.

\subsection{Overview}

\begin{figure*}[t]
  \centering
  \includegraphics[width=0.83\linewidth]{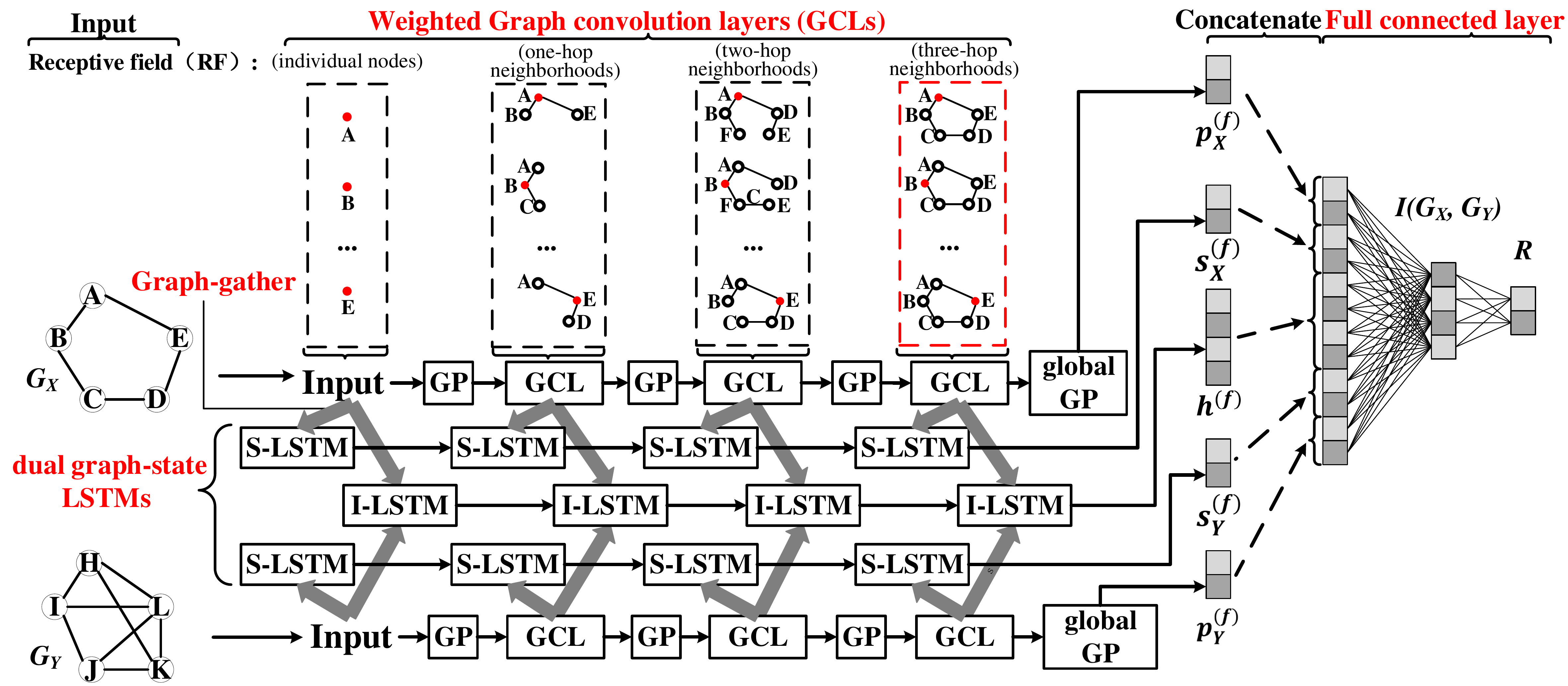}
  \caption{ A three-layer framework of MR-GNN.
    For each input graph, it uses several \textit{graph convolution layers}
    (\textbf{GCL}s) to learn multi-resolution structure features.
    Then, for each GCL, a \textit{graph-gather layer} sums the node vectors of the
    same resolution to get a graph-state.
    We feed the graph-states of different GCLs, which have different receptive
    fields, into our \textit{S-LSTM} and \textit{I-LSTM} to learn the final
    representation comprehensively.
    Finally, the final S-LSTM hidden vectors $s_{X}^{(f)}$ and $s_{Y}^{(f)}$, the
    final I-LSTM hidden vectors $h^{(f)}$, and the graph pooling (GP) vectors of
    entire graph $p_{X}^{(f)}$ and $p_{Y}^{(f)}$ are concatenated and passed to
    the following fully connected layers for learning a predictive model.
  }
  \label{fig:MR-GNN}
\end{figure*}

Figure~\ref{fig:MR-GNN} depicts the architecture of MR-GNN, which mainly consists
of three parts: 1) multiple {\em weighted graph convolution layers}, which extract
structure features from receptive fields of different sizes, 2) {\em dual
  graph-state LSTMs}, which summarize multi-resolution structure features and
extract interaction features, and 3) {\em fully connected layers}, which predict
the entity interaction labels.

\subsection{Weighted graph convolution layers}

Before introducing the motivation and design of our weighted graph convolution
operators in detail, we elaborate the standard graph convolution operator.

\noindent\textbf{Standard Graph Convolution Operator.}
Inspired by the convolution operator on images, for a specific node in a graph,
the general spatial graph convolution~\cite{Duvenaud2015} aggregates features of a
node as well as its one-hop neighbors' as the node's new features.
Based on the above definition, take the node $v_i$ as an example, the formula is:
\begin{equation}\label{eq:nbr_conv}
  f_i^{(t+1)}=
  \sigma\left( (f_i^{(t)} +  \sum_{v_j\in N_i} f_j^{(t)})W_{d_i}^{(t)} \right), t = 0, 1, \ldots
\end{equation}
where $f_i^{(t+1)}$ denotes the feature vector of $v_i$ in the $(t+1)^{\text{th}}$
graph convolution layer, $W_{d_i}^{(t)}$ is the weight matrix associated with the
center node $v_i$ and $\sigma(\cdot)$ is the tanh activation function.
Note that $f_i^{(0)}=f_i$.


Because the output graph of each graph convolution layer is exactly same as the input graph, MR-GNN can conveniently
  learn the structural characteristics of different resolutions through different
  iterations of the graph convolution layer.
Take the \textit{node A} in Fig.~3 as an example, after three iterations of graph convolution layer, the
receptive field in the third graph convolution layer is a
three-hop neighborhood centered on it.


However, since graphs are not regular grids compared with images, it is difficult
for the existing graph convolution operator to distinguish the weight by spatial
orientation position like the convolution operator on grid-like data,
\textit{e.g., in the image processing, the right neighbor and the left neighbor of
  a pixel can be treated with different weight for each convolution kernel.}
Inspired by the fact that the degree of nodes can well reflect the importance of
nodes in a network for many applications.
We modify the graph convolution operator by adding weights according to the node
degree $d_i$.
(\textit{Other metrics such as betweenness centrality can also work well.
  In this paper we choose the degree of nodes because of the simplicity of
  calculation.})
{Furthermore, Sukhbaatar et al.~\shortcite{sukhbaatar2016learning}
  treats different agents with different weights in order to distinguish the
  feature of the original node and the features of neighboring nodes.
  We treat each node and its neighbors with different weight matrixes, $\Phi$ and
  $\Psi$.}
Our improved weighted graph convolution is as follows:
\begin{equation}
  f_i^{\prime(t+1)}=
  f_i^{(t)}\Phi_{d_i}^{(t)}+\sum_{v_j\in N_i}f_j^{(t)}\Psi_{d_i}^{(t)}+b_{d_i}^{(t)}
\end{equation}
where ${\Phi_{d_i}^{(t)}, \Psi_{d_i}^{(t)}\in \mathbb{R}^{c_t \times c_{t+1}}}$
denote the weight of node $v_i$ with degree $d_i$, $c_{t+1}$
denotes the dimension of the feature vector in the $(t+1)^{\text{th}}$
graph convolution layer, and $b^{(t)}\in \mathbb{R}^{1 \times c_{t+1}}$ is a bias.
We let $c_0=c$.

After each convolution operation, similar to the classical CNN, we use a graph
pooling operation $GP(\cdot)$ to summarize the information within neighborhoods
(i.e., a center node and its neighbors).
For a specific node, the \textit{Graph Pooling}~\cite{altae2017low} returns a new feature vector
of which each element is the maximum activation of the corresponding element of one-hop neighborhood at this node.
We denote this operation by the following formula and get the feature vectors of the next
layer:
\begin{equation}
  f_i^{(t+1)} = GP(f_i^{\prime(t+1)},\{f_j^{\prime(t+1)}\}_{v_j \in N_i})
\end{equation}


\subsection{Graph-gather layers}
Graph interaction prediction is a graph-level problem rather than a node-level
problem.
To learn the graph-level features of different-sized receptive fields, we
aggregate the node representations of each convolution layer's graph to a
graph-state by a \textbf{graph-gather} layer.
Graph-gather layers compute a weighted sum of all node vectors in the connected
graph convolution layers.
The formula is:
\begin{equation}
  g^{(t)} = \sum_{1\leq i \leq m}f_i^{(t)} \Theta^{(t)}_{d_i}+\beta_{d_i}^{(t)}
\end{equation}
where $\Theta^{(t)}_{d_i}\in \mathbb{R}^{c_t \times c_{G}}$ is the graph-gather
weight of nodes with $d_i$ degree in the $t^{\text{th}}$ graph convolution
layer, $g^{(t)}$ is the graph-state vector of the $t^{th}$
convolution layer, $c_{G}$ denotes the dimension of graph-states, $m$ is the nodes' number
in the graph and $\beta_{d_i}^{(t)}\in \mathbb{R}^{1 \times c_{G}}$ is a bias.
Specially, the first graph-state $g^{(0)}$ only includes all individual nodes' information.


\subsection{Dual graph-state lstms}
To solve graph-level tasks, the existing graph convolution networks (GCNs)
methods~\cite{altae2017low} generally choose the graph-state of the last
convolution layer, which has the largest receptive fields, as input for subsequent
prediction.
But such state may loss many important features.

Referring to the CNN on images, there are multiple convolution kernels for
extracting different features in each convolution layer, which ensure the hidden
representation of the final convolution layer can fully learn features of input
images.
However, GCN is equivalent to CNN that only has one kernel in each layer.
It is difficult for the output of the final graph convolution layer to fully learn
all features in the large receptive fields, especially for structure features of
small receptive field.
The straightforward way is to design multiple graph convolution kernels and
aggregate the output of them.
However it is computational expensive.

To solve the above problem, we propose a multi-resolution based architecture in our model, in which the graph-state of each graph convolution layer is leveraged to learn the final representation.
We propose a Summary-LSMT (S-LSTM) to aggregate the
graph-states of different-sized receptive fields for learning the final features
comprehensively.
Instead of the straightforward method that directly sums all graph-states up,
S-LSTM models the node information diffusion process of graphs by sequentially
receiving the graph-state $g^{(t)}$ with receptive field from small to large
as inputs.
It is inspired by the idea \textit{a representation that encapsulates graph
  diffusion can provide a better basis for prediction than the graph itself}.
The formula of S-LSTM is:
\begin{equation}
  s^{(t+1)} = \mathrm{LSTM}(s^{(t)}, g^{(t)})
\end{equation}
where $s^{(t+1)}\in \mathbb{R}^{1 \times c_{G}}$ is the ${(t+1)}^{th}$ hidden
vector of S-LSTM.
To further enhance the global information of graphs, we concatenate the final
hidden output $S^{(f)}$ of S-LSTM and the output $p^{(f)}$ of global graph pooling
layer as the final graph-state of the input graph:
\begin{equation}
  e^{(f)} = [s^{(f)},p^{(f)}]
\end{equation}
where $p^{(f)}=GP(f_{v_1}^{(f)},...,f_{v_m}^{(f)})\in\mathbb{R}^{1\times c_f}$ is
the result of global graph pooling on the final graph convolution layer.

In addition, to extract the interaction features of pairwise graphs, we propose an
Interaction-LSTM (I-LSTM) which takes the concatenation of dual graph-states as
input:
\begin{equation}
  h^{(t+1)} = \mathrm{LSTM}(h^{(t)}, [g_X^{(t)},g_Y^{(t)}])
\end{equation}
where $h^{(t+1)}\in \mathbb{R}^{1 \times 2c_{G}}$ is the ${(t+1)}^{th}$ hidden
vector of I-LSTM .We initialize $s^{(0)}$ and $h^{(0)}$as an all-zero vector and
the S-LSTM is shared to both input graphs.

\subsection{Fully connected layers}
For the interaction prediction, we simply concatenate the final graph
representations and interaction features of input graphs (i.e., $e_X^{(f)}$,
$e_Y^{(f)}$ and $h^{(f)}$) and use fully connected layers for prediction.
Formally, we have:
\begin{align}
  & I(G_X,G_Y) = [e_X^{(f)}, e_Y^{(f)}, h^{(f)}] \\
  & R = \sigma_s(f_2(\sigma_r(f_1(I(G_X,G_Y)))))
\end{align}
where $f_i(x)= W_ix+b_i, i=1,2,$ are linear operations, $W_1\in
\mathbb{R}^{(2c_{f}+ 4c_{G}) \times c_k}$ and $W_2\in \mathbb{R}^{c_k \times k}$
are trainable weight matrices, $c_k$ is the dimension of the hidden vector, and
$k$ is the number of interaction labels.
The activation function $\sigma_r(\cdot)$ is a rectified linear unit (ReLU), i.e.,
$\sigma_r(x) = max(0,x)$.
$R$ is the output of softmax function $\sigma_s(\cdot)$, the $j^{th}$ element of
$R$ is computed as $r_j = \frac{e^{r_j}}{\sum_{i=0}^k e^{r_i}}$.
At last, we choose the cross entropy function as loss function, that is:
\begin{equation}
  L(R, \hat{R}) = -\sum_{i=1}^k \hat{r_i}log(r_i)
\end{equation}
where $\hat{R}\in \mathbb{R}^{1 \times k}$ is the ground-truth vector.

\section{Experiment}\label{Experiment}

In this section, we conduct experiments to validate our method\footnote{Code available at \url{https://github.com/prometheusXN/MR-GNN}}.
We consider two prediction tasks: 1) predicting whether there is an interaction
between two chemicals (i.e., binary classification), and 2) predicting the
interaction label between two drugs (i.e., multi-class classification).

\subsection{Dataset}

\noindent\textbf{CCI Dataset.}
For the binary classification task, we use the CCI
dataset\footnote{\url{http://stitch.embl.de/download/chemical_chemical.links.detailed.v5.0.tsv.gz}}.
This dataset uses a score ranging from $0$ to $999$ to describe the interaction
level between two compounds.
The higher the score is, the larger probability the interaction will occur with.
According to threshold scores $900$, $800$ and $700$, we got positive samples of
three datasets: CCI$900$, CCI$800$, and CCI$700$.
As for negative samples, we choose the chemical pairs of which the score is $0$.
For each pair of chemicals, we assign a label ``1" or ``0" to indicate whether an
interaction occurs between them.
We use a public available API, DeepChem\footnote{\url{https://deepchem.io/}}, to
convert compounds to graphs, that each node has a 75-dimension feature vector.

\noindent\textbf{DDI Dataset.}
For the multi-class classification task, we use the DDI
dataset\footnote{\url{http://www.pnas.org/content/suppl/2018/04/14/1803294115.DCSupplemental}}.
This dataset contains $86$ interaction labels, and each drug is represented by
SMILES string~\cite{Weininger1988}.
In our preprocessing, we remove the data items that cannot be converted into
graphs from SMILES strings.


\begin{table}[htp]
\small
\centering
\caption{Statistics of datasets.}
\label{tab:data}
\begin{tabular}{@{}cccc@{}}
\toprule
Dataset & Graph Meaning        & \#Graphs & \#Pairs \\
\midrule
CCI900  & Chemical Compounds   & 11990    & 19624   \\
CCI800  & Chemical Compounds   & 73602    & 151796  \\
CCI700  & Chemical Compounds   & 114734   & 343277  \\
DDI     & Drug Molecule Graphs & 1704     & 191400  \\
\bottomrule
\end{tabular}
\end{table}

\subsection{Baselines}

\begin{table*}[t]
\small
\centering
\caption{Experimental results of the binary classification task.}
\label{tab:binary}
\begin{tabular}{@{}l@{ }|@{ }c@{\ \ }c@{\ \ }c@{\ \ }c@{ }|@{ }c@{\ \ }c@{\ \ }c@{\ \ }c@{ }|@{ }c@{\ \ }c@{\ \ }c@{\ \ }c@{}}
\hline
                & \multicolumn{4}{c|}{CCI900} & \multicolumn{4}{c|}{CCI800} & \multicolumn{4}{c}{CCI700} \\
\hline
        & AUC           & accuracy      & recall        & F1            & AUC    &accuracy  & recall  & F1      & AUC     & accuracy&recall   & F1   \\
\hline
PIP     & $93.92$ & $87.95$ & $88.73$ & $87.66$ & $98.49$ & $94.67$ & $94.74$ & $94.59$ & $98.92$ & $95.53$ & $94.96$ & $95.52$ \\
SNR     & $91.86$ & $83.99$ & $79.40$ & $82.95$ & $97.18$ & $91.19$ & $89.81$ & $90.95$ & $98.04$ & $92.87$ & $92.21$ & $92.85$ \\
DGCNN   & $95.14$ & $85.53$ & $84.72$ & $85.12$ & $97.13$ & $91.54$ & $91.55$ & $91.43$ & $97.95$ & $93.13$ & $92.65$ & $93.13$ \\
DeepDDI & $90.30$ & $83.74$ & $82.94$ & $83.58$ & $95.43$ & $89.73$ & $90.10$ & $89.88$ & $96.48$ & $91.77$ & $91.74$ & $91.80$ \\
DeepCCI & $95.14$ & $88.11$ & $88.90$ & $87.95$ & $98.69$ & $95.38$ & $94.93$ & $95.34$ & $99.22$ & $96.25$ & $95.54$ & $96.25$ \\
{\bf MR-GNN} & $\bf 95.67$ & $\bf 90.16$ & $\bf 91.21$ & $\bf 90.05$ & $\bf 98.76$ & $\bf 95.44$ & $\bf 95.28$ & $\bf 95.38$ & $\bf 99.25$ & $\bf 96.51$ & $\bf 96.08$ & $\bf 96.51$ \\
\hline
\end{tabular}
\end{table*}

We compare our method with the following state-of-the-art models:
\begin{itemize}
\item \textbf{DeepCCI}
    ~\cite{Kwon2017} is one of the
  state-of-the-art methods on the CCI datasets.
  It represents SMILES strings of chemicals as one-hot vector matrices and use
  classical CNN to predict interaction labels.

\item \textbf{DeepDDI}~\cite{DeepDDI} is one of the state-of-the-art methods on
  the DDI dataset.
  DeepDDI designs a feature called structural similarity profile (SSP) combined
  with multilayer perceptron (MLP) for prediction.
\end{itemize}


\begin{itemize}
\item \textbf{PIP} ~\cite{protein_interface} is proposed to predict the protein
  interface.
  It extracts features from the fixed three-hop neighborhood for each node to
  learn a node representation.
  In this paper, when building this model, we use our graph-gather layer to
  aggregate node representations to get the graph representation.

\item \textbf{DGCNN}~\cite{DGCNN} uses the standard graph convolution operator as
  described in Section 3.
  It concatenates the node vectors of each graph convolution layer and applies CNN
  with a node ordering scheme to generate a graph representation.

\item \textbf{SNR}~\cite{Li2017} uses the similar graph convolution layer as our
  method.
  The difference is that this work introduces an additional node that sums all
  nodes features up to a graph representation.
\end{itemize}

\subsection{Binary classification}

\noindent\textbf{Settings.}
We divide each CCI dataset into a training dataset and a testing dataset with
ratio $9:1$, and randomly choose $1/5$ of the training dataset as a validation
dataset.
We set the three graph convolution layers with $384$, $384$, $384$ output units,
respectively.
We set $128$ output units of graph-gather layers as the same as the LSTM layer.
The fully connected layer has $64$ hidden units followed by a softmax layer as the
output layer.
We set the learning rate to $0.0001$.
To evaluate the experimental results, we choose four metrics: \emph{area under ROC
  curve (AUC), accuracy, recall, and F1}.

\noindent\textbf{Results.}
Table~\ref{tab:binary} shows the performance of different methods.
MR-GNN performs the best in terms of all of the evaluation metrics.
Compared with the state-of-the-art method DeepCCI, our MR-GNN improves accuracy by $0.6\%$-$2.5\%$, F1 by $0.6\%$-$2.2\%$, recall by $1.0\%$-$2.3\%$, and AUC by
$0.05\%$-$0.32\%$.
As for little improvement of AUC, we think it is ascribed to the fact that the
basic value is too large to provide enough space for improvement.
When translated into the remaining space, the AUC is increased by
$4.2\%$-$11.8\%$.
The performance improvement proves that features extraction of MR-GNN, which
represents structured entities as graphs for features extraction, is more
effective than DeepCCI, which treats SMILES string as character sequence without
considering topological information of structured entities.
Compared with PIP, the performance of MR-GNN demonstrates that the
multi-resolution based architecture is more effective than the fix-sized RF based
framework.
In addition, compared with SNR which directly sums all node features to get the
graph representation, experimental results prove that our S-LSTM summarizes the
local features more effectively and more comprehensively.
We attribute this improvement to the diffusion process and the interaction that
our graph-state LSTM modeled during the procedure of feature extraction, which is
effective for the prediction.

\subsection{Multi-class classification}

{
\noindent\textbf{Settings.}
To make an intuitional comparison, similar to DeepDDI, we use $60\%$, $20\%$,
$20\%$ of dataset for the training, validation and testing, respectively.
All hyper-parameter selections are the same as the binary classification task.
To evaluate the experimental results, we choose five metrics on the
multi-classification problem: {\em AUPRC}, {\em Micro average}, {\em Macro
  recall}, {\em Macro precision}, and {\em Macro F1}.
(In particular, we choose the AUPRC metric due to the imbalance of the DDI
dataset.)
We show the results on DDI dataset in Table~\ref{tab:multi}.

\begin{table}[bp]
\footnotesize
\centering
\caption{Results on the DDI dataset.}
\label{tab:multi}
\begin{tabular}{@{}l@{ }|@{\ }c@{\ \ }c@{\ \ }c@{\ \ }c@{\ \ }c@{}}
\hline
                & Mi\_avg        & Ma\_recall     & Ma\_pre        & Ma\_F1      & AUPRC       \\
\hline
PIP             & $92.73$        & $87.45$        & $89.47$        & $87.88$     & $91.60$     \\
SNR             & $82.91$        & $79.88$        & $76.71$        & $76.88$     & $81.85$     \\
DGCNN           & $86.63$        & $72.02$        & $79.15$        & $74.32$     & $85.54$     \\
DeepCCI         & $87.38$        & $79.91$        & $88.86$        & $82.73$     & $86.29$     \\
DeepDDI         & $92.64$        & $83.86$        & $89.58$        & $85.70$     & $91.52$     \\
\hline
{\bf MR-GNN}    & $\bf 94.31$    & $\bf 92.68$    & $\bf 94.94$    & $\bf 93.48$ & $\bf 93.18$ \\
\hline
{\bf -no I-LSTM} & $\bf 94.11$    & $\bf 91.96$    & $\bf 94.23$    & $\bf 92.81$ & $\bf 92.98$ \\
{\bf -no S-LSTM} & $\bf 94.05$    & $\bf 90.31$    & $\bf 94.53$    & $\bf 91.82$ & $\bf 92.92$ \\
{\bf -no w-GCL}  & $\bf 93.86$    & $\bf 89.33$    & $\bf 92.83$    & $\bf 90.38$ & $\bf 92.74$ \\
{\bf -no LSTMs}  & $\bf 92.83$    & $\bf 86.38$    & $\bf 91.61$    & $\bf 88.11$ & $\bf 91.71$ \\
\hline
\end{tabular}
\end{table}

\noindent\textbf{Results.}
We observe that MR-GNN performs the best in terms of all five evaluation metrics.
MR-GNN improves these five metrics by $1.58\%$, $5.23\%$, $5.46\%$, $5.60\%$ and $
1.58\%$, respectively.
Compared with the state-of-the-art method DeepDDI, the performance improvement of
MR-GNN is attributed to the higher quality representations learned by end-to-end
training instead of the human-designed representation called SSP.
In addition, we also conduct experiments on CCI and DDI datasets, and we observe
that MR-GNN indeed improves performance.

\noindent\textbf{Ablation experiment.}
We also conducted ablation experiments on the DDI dataset to study the effects of
three components in our model (namely S-LSTM, I-LSTM, and weighted GCL).
We find that each of these three components can improve performance.
Among them, weighted GCLs contributes most significantly, then comes S-LSTM and
I-LSTM.

}

\subsection{Efficiency and robustness}

In the third experiment, we conduct experiments to analyze the efficiency and
robustness of MR-GNN.

\begin{figure}[t]
  \centering
  \subfigure[\label{f:cci900_accuracy}]{%
   \includegraphics[width=0.46\linewidth]{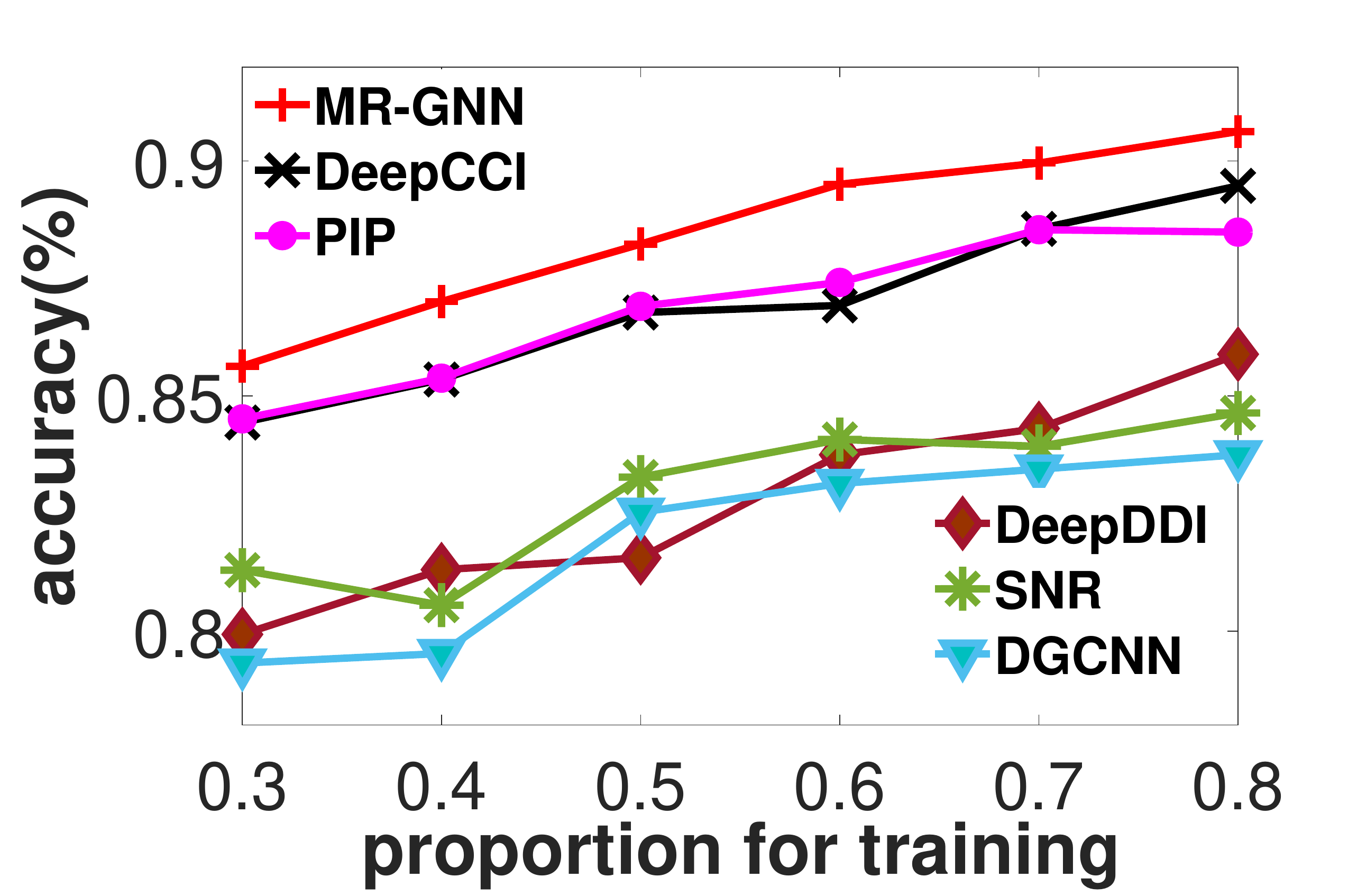}}
  \subfigure[\label{f:cci900_time}]{%
  \includegraphics[width=0.46\linewidth]{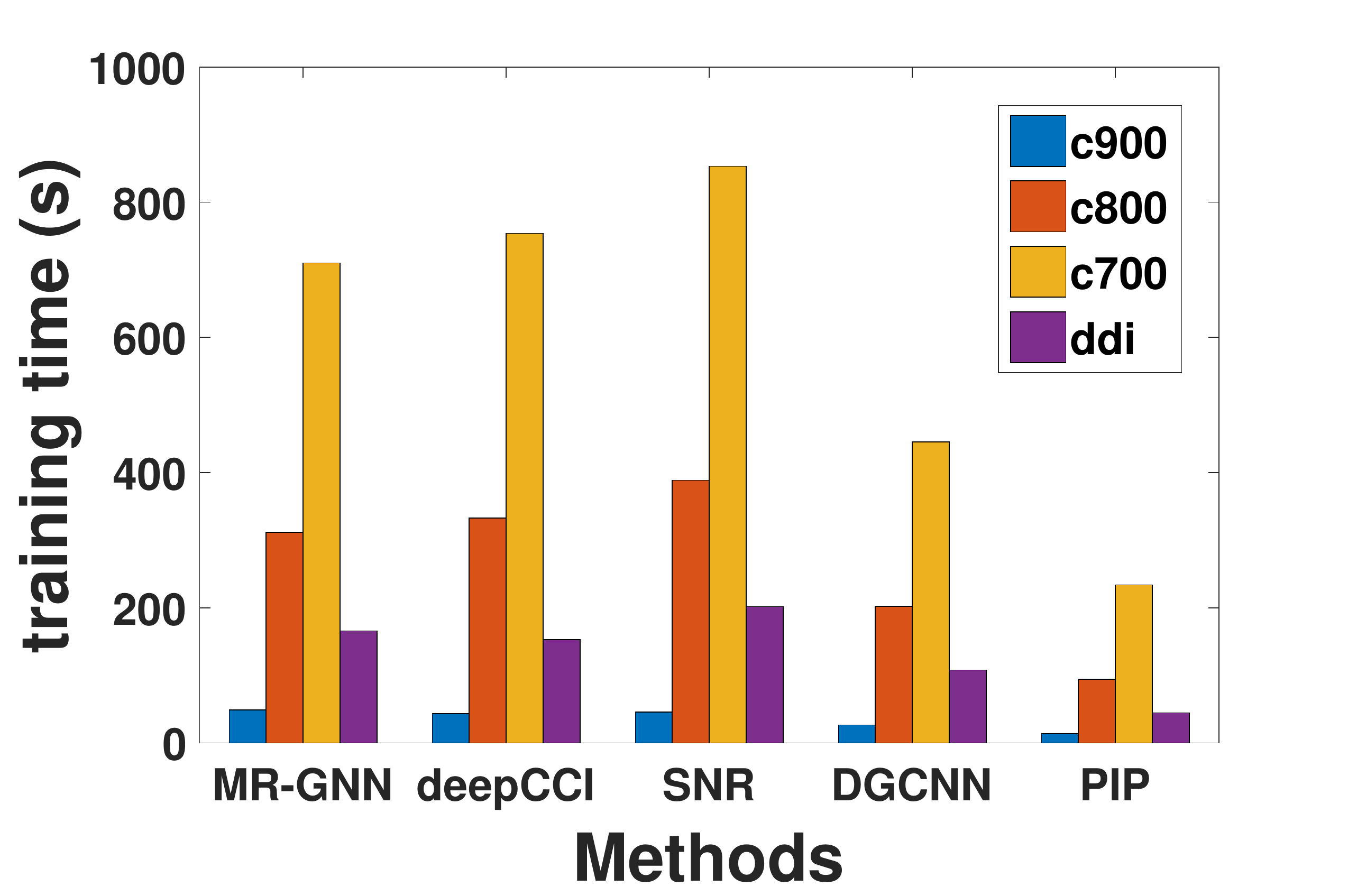}}
  \caption{Result on CCI900:
    a) Accuracy under different training set proportions;
    b) Training time per epoches.
  }
\end{figure}

\noindent\textbf{Effects of training dataset size.}
We carried out a comparative experiment with different size of training datasets
from $30\%$ to $70\%$ on the CCI900 dataset.
In each comparative experiments, we kept the same $10\%$ of the dataset as the
test dataset to evaluate the performance of all six methods.
Figure~\ref{f:cci900_accuracy} shows that MR-GNN always performs the best under
different training dataset size.
In particular, as the training dataset proportions increases, the improvement of
MR-GNN increases significantly, demonstrating that our MR-GNN has better
robustness.
This is due to the fact that MR-GNN is good at learning subgraph information of
different-sized receptive fields, especially subgraphs of small receptive fields
that often appear in various graphs.

\noindent\textbf{Training efficiency.}
Figure~\ref{f:cci900_time} shows that the training time of MR-GNN is at a moderate
level among all methods.
Although the graph-state LSTMs takes the additional time, the training of MR-GNN
is still fast and acceptable.

\begin{figure}[t]
  \centering
  \subfigure[\label{f:represent_size}]{%
  \includegraphics[width=0.46\linewidth]{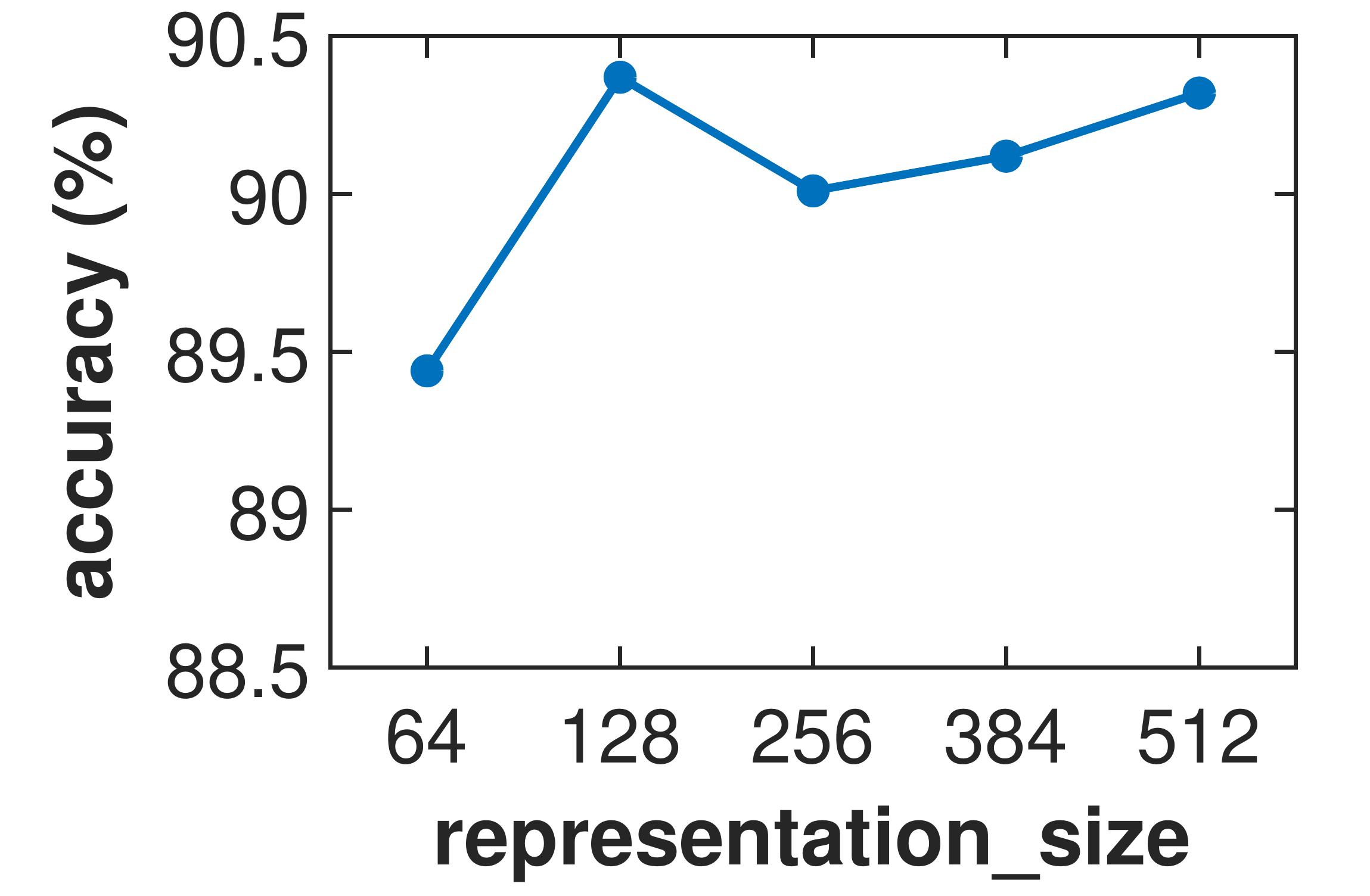}}
  \subfigure[\label{f:conv_size}]{%
  \includegraphics[width=0.46\linewidth]{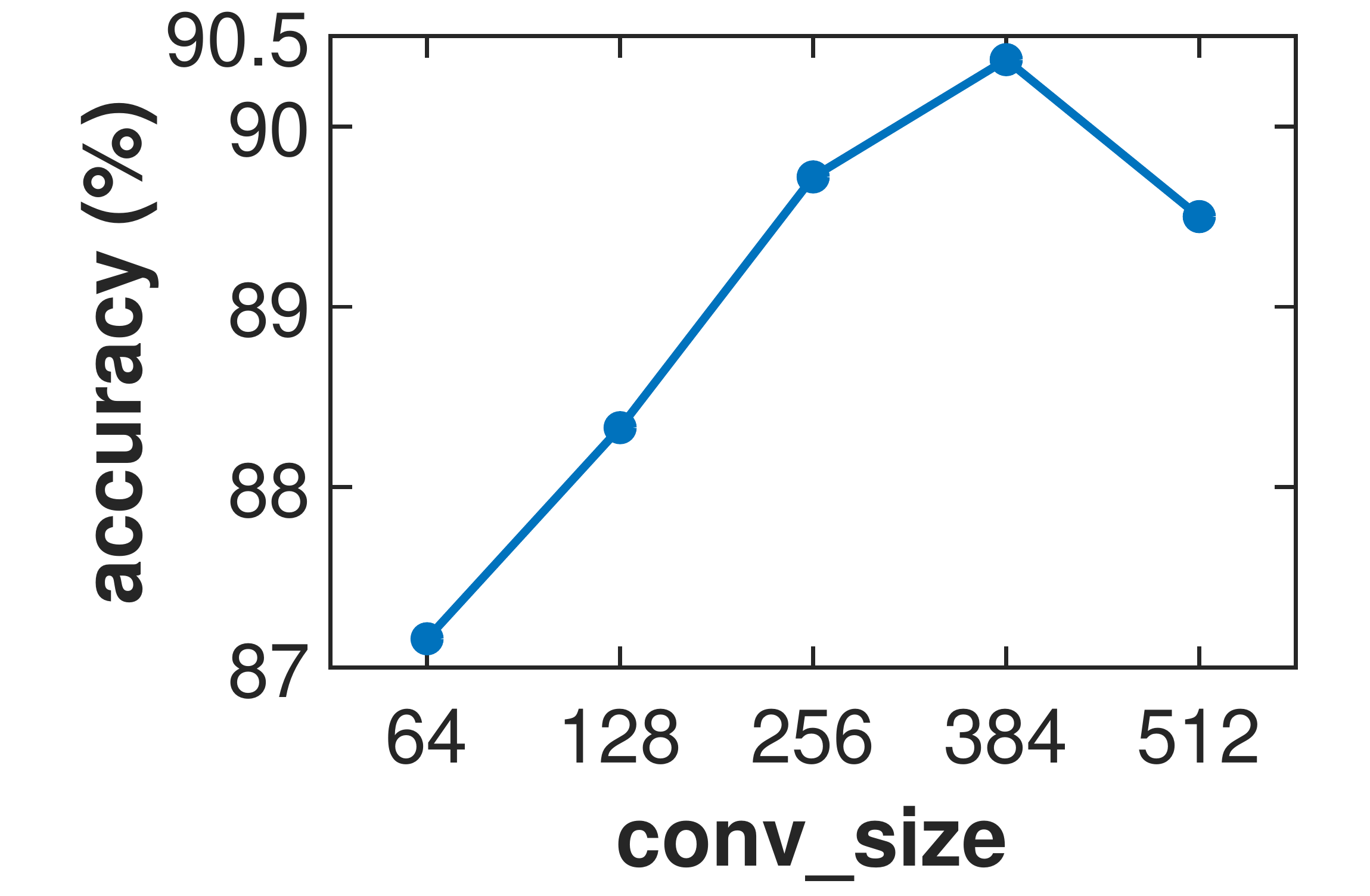}}
  \subfigure[\label{f:output_size}]{%
  \includegraphics[width=0.46\linewidth]{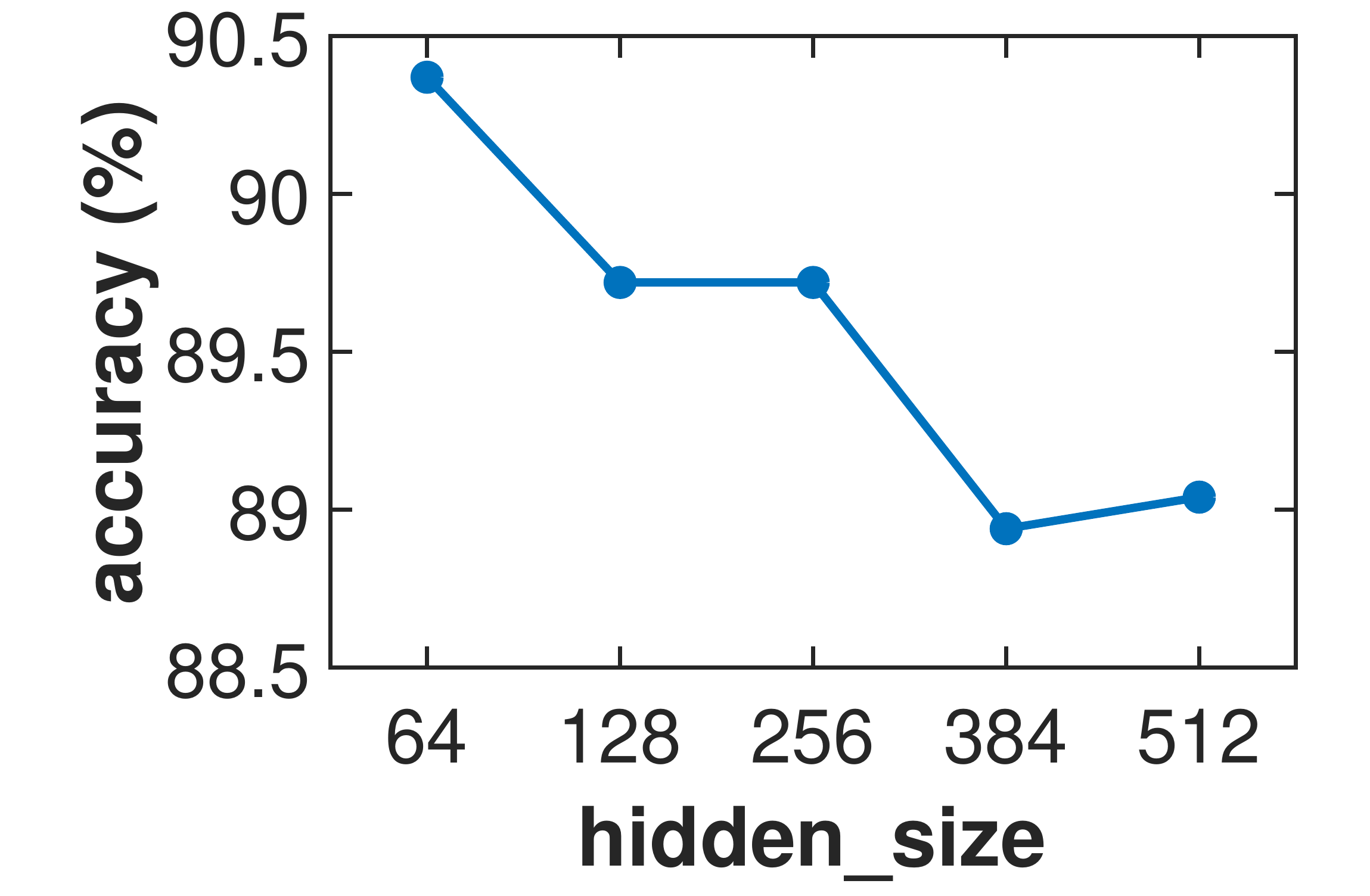}}
  \subfigure[\label{f:learning_rate}]{%
  \includegraphics[width=0.46\linewidth]{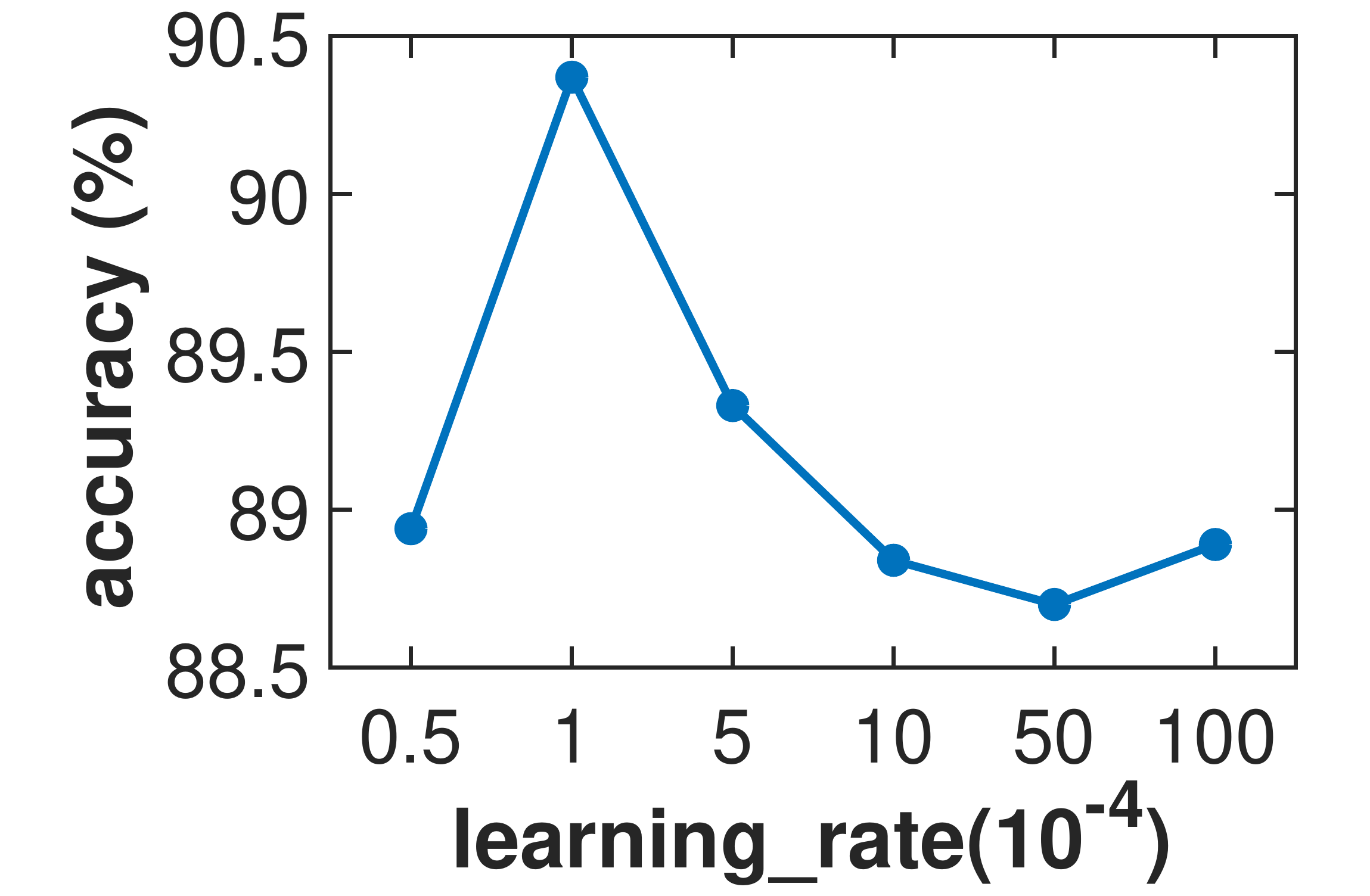}}
\caption{Parameter sensitivities w.r.t.~$represent\_size$, $conv\_size$,
  $hidden\_size$ and $learning\_rate$.}
\label{fig:parameters}
\end{figure}

\noindent\textbf{Effects of hyper-parameter variation.}
In this experiment, we consider the impact of hyper-parameters of MR-GNN: the
output units number of GCLs ($conv\_size$) and LSTMs
($represent\_size$), the hidden units number of the fully connected layer
($hidden\_size$), and $learning\_rate$.
The results are shown in Fig.~\ref{fig:parameters}.
We see that the impact of hyper-parameter variation is insignificant (the absolute
difference is less than $2\%$).
Fig.~\ref{f:represent_size} shows that larger $represent\_size$ provides a better
performance (with an salient point at $represent\_size = 128$).
Fig.~\ref{f:conv_size} shows that similar result of $conv\_size$ while a salient point is at $conv\_size = 384$.
The performance increases fast when $conv\_size < 384$ and slightly declines when
$conv\_size \ge 384$.
As for $learning\ rate$ and $hidden\_size$, the best point appears at $1\times10^{-4}$ and $512$,
respectively.

\section{Related Work}\label{related}

\noindent{\textbf{Node-level Applications.}}
Many neural network based methods have been proposed to solve the node-level tasks
such as \textit{node
  classification}~\cite{henaff2015deep,li2015gated,defferrard2016convolutional,Kipf2016Semi,attenteion_representation},
\textit{link prediction}~\cite{link_prediction,zhang2018graph}, etc.
They rely on node embedding techniques, including skip-gram based methods like
DeepWalk~\cite{Deepwalk} and LINE~\cite{Line}, autoencoder based methods like
SDNE~\cite{SDNE}, neighbor aggregation based methods like
GCN~\cite{defferrard2016convolutional,Kipf2017} and GraphSAGE~\cite{GraphSAGE},
etc.

\noindent{\textbf{Single Graph Based Applications.}}
Attention also has been paid on the graph-level tasks.
Most existing works focus on classifying graphs and predicting graphs'
properties~\cite{Duvenaud2015,atwood2016diffusion,Li2017,DGCNN} and they compute
one embedding per graph.
To learn graph representations, the most straightforward way is to aggregate node
embeddings, including average-based methods (simple average and weight average)
~\cite{Li2017,Duvenaud2015,zhao2018substructure}, sum-based
methods~\cite{sum_based_method} and some more sophisticated schemes, such as
aggregating nodes via histograms~\cite{Kearnes2016Molecular} or learning
node ordering to make graphs suitable for CNN~\cite{DGCNN}.

\noindent{\textbf{Pairwise Graph Based Applications.}}
Nowadays, very little neural network based works pay attention to the pairwise
graph based tasks whose input is a pair of graphs.
However, most existing works focus on learning ``similarity" relation between
graphs~\cite{GraphEdit,Kernel2015Deep} or links between nodes across
graphs~\cite{protein_interface}.
In this work, we study the prediction of the universal graph interactions.

\section{Conclusion}\label{conclusion}

In this paper, we propose a novel graph neural network, i.e., MR-GNN, to predict
the interactions between structured entities.
MR-GNN can learn comprehensive and effective features by leveraging a
multi-resolution architecture.
We empirically analyze the performance of MR-GNN on different interaction
prediction tasks, and the results demonstrate the effectiveness of our model.
Moreover, MR-GNN can easily be extended to large graphs by assigning node weights
to node groups that based on the distribution of node degrees.
In the future, we will apply it to more other domains.

\section*{Acknowledgments}
The research presented in this paper is supported in part by National Key R\&D Program of China
(2018YFC0830500), National Natural Science Foundation of China (UI736205, 61603290),
Shenzhen Basic Research Grant (ICYJ20170816100819428),
Natural Science Basic Research Plan in Shaanxi Province of China (2019JM-159),
Natural Science Basic Research Plan in ZheJiang Province of China (LGG18F020016).

\bibliographystyle{named}
\bibliography{ref}

\end{document}